\documentclass[a4paper, 10pt, conference]{IEEEtran}
\IEEEoverridecommandlockouts
\usepackage{cite}
\usepackage{multirow}
\usepackage{amsmath,amssymb,amsfonts}
\usepackage{algorithmic}
\usepackage{graphicx}
\usepackage{textcomp}
\usepackage{xcolor}
\usepackage{array}
\usepackage{eso-pic}
\usepackage{lipsum}

\def\BibTeX{{\rm B\kern-.05em{\sc i\kern-.025em b}\kern-.08em
 T\kern-.1667em\lower.7ex\hbox{E}\kern-.125emX}}
\begin{document}

\AddToShipoutPictureBG*{%
  \AtPageUpperLeft{%
    \hspace{\paperwidth}%
    \raisebox{-\baselineskip}{%
      \makebox[0pt][r]{\tiny{\textcopyright 2022 IEEE. Personal use of this material is permitted. Permission from IEEE must be obtained for all other uses, in any current or future media, including reprinting/republishing this material for advertising or promotional purposes, creating new collective works, for resale or redistribution }}
}}}%
\AddToShipoutPictureBG*{%
  \AtPageUpperLeft{%
    \hspace{\paperwidth}%
    \raisebox{-1.5\baselineskip}{%
      \makebox[0pt][r]{\tiny{to servers or lists, or reuse of any copyrighted component of this work in other works.} }
}}}%

\title{HyperDog: An Open-Source Quadruped Robot Platform Based on ROS2 and micro-ROS\\
}

\author{ 
 \IEEEauthorblockN{Nipun Dhananjaya Weerakkodi\\ Mudalige}
 \IEEEauthorblockA{\textit{Intelligent Space Robotics Laboratory} \\
 \textit{Skoltech}\\
 Moscow, Russian Federation \\
 nipun.weerakkodi@skoltech.ru}
\and
 \IEEEauthorblockN{Iana Zhura}
 \IEEEauthorblockA{\textit{Intelligent Space Robotics Laboratory} \\
 \textit{Skoltech}\\
 Moscow, Russian Federation \\
 iana.zhura@skoltech.ru}
\and
 \IEEEauthorblockN{Ildar Babataev}
 \IEEEauthorblockA{\textit{Intelligent Space Robotics Laboratory} \\
 \textit{Skoltech}\\
 Moscow, Russian Federation \\
 ildar.babataev@skoltech.ru}
\and
 \IEEEauthorblockN{Elena Nazarova}
 \IEEEauthorblockA{\textit{Intelligent Space Robotics Laboratory} \\
 \textit{Skoltech}\\
 Moscow, Russian Federation \\
 elena.nazarova@skoltech.ru}
\and
 \IEEEauthorblockN{Aleksey Fedoseev}
 \IEEEauthorblockA{\textit{Intelligent Space Robotics Laboratory} \\
 \textit{Skoltech}\\
 Moscow, Russian Federation \\
 aleksey.fedoseev@skoltech.ru}
\and
 \IEEEauthorblockN{Dzmitry Tsetserukou}
 \IEEEauthorblockA{\textit{Intelligent Space Robotics Laboratory} \\
 \textit{Skoltech}\\
 Moscow, Russian Federation \\
 d.tsetserukou@skoltech.ru}
}

\maketitle

\begin{abstract}
Nowadays, design and development of legged quadruped robots is a quite active area of scientific research. In fact, the legged robots have become popular due to their capabilities to adapt to harsh terrains and diverse environmental conditions in comparison to other mobile robots. With the higher demand for legged robot experiments, more researches and engineers need an affordable and quick way of locomotion algorithm development. In this paper, we present a new open source quadruped robot HyperDog platform, which features 12 RC servo motors, onboard NVIDIA Jetson nano computer and STM32F4 Discovery board. HyperDog is an open-source platform for quadruped robotic software development, which is based on Robot Operating System 2 (ROS2) and micro-ROS. Moreover, the HyperDog is a quadrupedal robotic dog entirely built from 3D printed parts and carbon fiber, which allows the robot to have light weight and good strength. The idea of this work is to demonstrate an affordable and customizable way of robot development and provide researches and engineers with the legged robot platform, where different algorithms can be tested and validated in simulation and real environment. The developed project with code is available on GitHub.

\footnote{{https://github.com/NDHANA94/hyperdog\_ros2}}

\end{abstract}


\section{Introduction}

Humans and animals have the ability to traverse rough and harsh terrains relying on their legged locomotion. With the advance in autonomous systems, legged robots inspired researchers and engineers to design and develop stable walking platforms for scenarios such as search and rescue, inspection, and exploration.
\begin{figure}[htbp]
\centering
\includegraphics[scale=0.28]{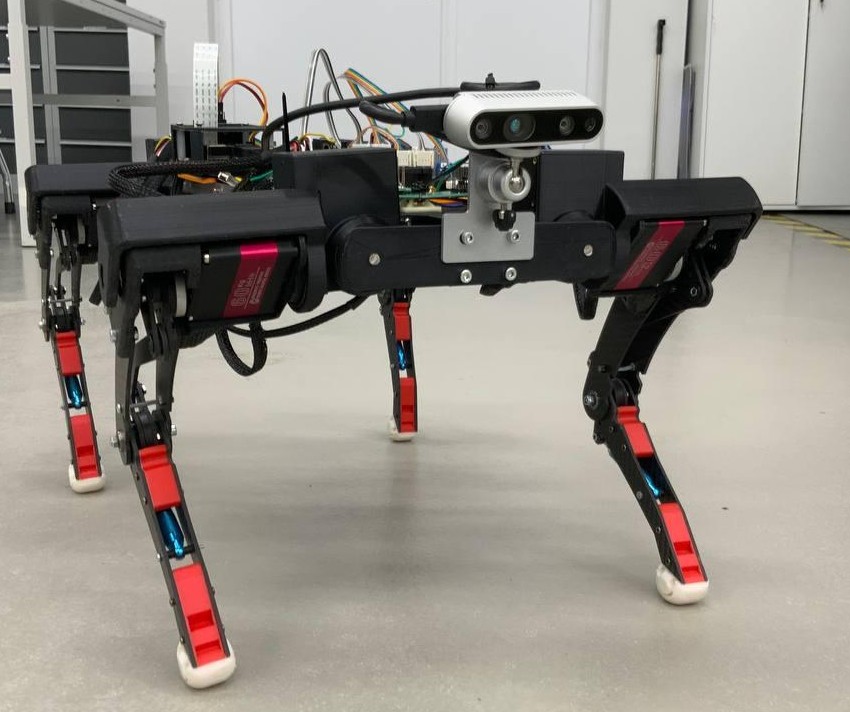}
\caption{HyperDog quadruped robot.}
\label{fig: Dog}
\end{figure}
For example, Raibert et al. \cite{BigDog} developed the BigDog quadruped robot demonstrating the ability to traverse rough terrains in 2008. However, the BigDog requires a high power density due to its hydraulic actuators and produces loud noise due to a two-stroke gasoline engine. Bellicoso et al. \cite{advances} evaluated the performance of the quadruped robot ANYmal in outdoor missions of industrial inspection and search and rescue competitions. Kim et al. \cite{Vision} integrated two Intel RealSense cameras into the MIT Mini-Cheetah and developed filtering algorithms for navigation on highly irregular terrain. The mentioned solutions achieved high performance on irregular terrains, yet they are still require further development and evaluations in multi-terrain environments, e.g., stairs, rivets, etc. There have been several approaches to scale down the hydraulically actuated quadruped robots, for example, MiniHyQ and HyQ2Max quadruped robots developed by Khan et al.\cite{MiniHyQ} and Semini et al. \cite{HyQ2Max}. However, hydraulic actuators have not yet been widely adopted for legged robots due to their noise, weight, cost, and maintenance complexity.

On the other hand, electric motors are a promising solution for quadruped robot design with less noise, low power density, and fewer maintenance issues. However, electric motors have lower torque and are not suitable for directly driving joints of legged robots. To overcome this problem, all the electric motors driving quadruped robots use high ratio gear reducers such as harmonic drives, planetary gear drives, etc, to achieve sufficient torque for the robot joints.
For example, Hutter et al. \cite{starleth} developed a kinematically controlled StarlETH serial elastic robot of 23 kg weight and 0.5 m tall, actuated by electric motors.
Katz et al. \cite{miniCheetah} introduced Mini-Cheetah that has advanced dynamic locomotion ability. One of the main novelties of this robot is the developed high-power modular electric actuators. The Mini-Cheetah has 12 identical actuators driven by brush-less direct current (BLDC) motors with planetary gears with a 6:1 reduction ratio, an encoder, and a motor controller. All of the actuators communicate over a single CAN bus. Another advanced dynamic quadruped robot is ANYmal by Hutter et al. \cite{ANYmal}. The robot design of 30 kg weight and 0.5 m height achieved a highly robust performance against impulsive loads during running and jumping. The advanced versions of the ANYmal robotic dog were tested and evaluated in several real-world applications researches\cite{fankhauser2018anymal}. 
However, the adaptation of quadruped robots to diverse terrains is still a significant challenge. Before real-life robot experiments, locomotion algorithms should be tested in simulations. Moreover, there are few developed open-source platforms, which can help to test the movements of legged robots, and these platforms do not support advanced locomotion algorithms.



\section{Related Works}
With a high demand for legged robot exploration of the environment, several open-source robotic platforms have appeared. For example, Vitruvio by Chadwick et al. \cite{Vitruvio} is the open-source leg design optimization toolbox for walking robots, which allows researchers to generate realistic leg motion and conduct analysis of the joint torque and mechanical energy. The key novelty of the framework is the computational optimization method applied in analyzing a robot design solution. A number of the open-source projects develop a hardware platforms for evaluation of the locomotion algorithms, e.g., the Stanford-Doggo by Kau et al. \cite{Doggo} proposing the torque-controlled legged platform. 
 
Several existing projects are focusing on the dynamic analysis of a robot leg design. An open-source kit for education with dynamic legged robot HOPPY was developed by Joao Ramos et al \cite{HOPPY}. HOPPY is a one-legged robot, which dynamically hops around a fixed gantry. Moreover, different quadrupedal designs have been explored, for example, Boxi Xia et al. proposed a real-time soft robot simulation environment for dynamic locomotion analysis \cite{Soft}. In this work, developed locomotion algorithms have been tested on a real quadrupedal robot traversing various terrains. To overcome these challenges, the soft robot has performed different gait patterns such as turning and back-flipping. 
 
Aside from the problems of obstacle avoidance and navigation on harsh terrains, the development of a high-velocity legged robot with light weight is a topic of active investigation by researchers. For example, the PAWDQ quadruped robot by Kim et al. \cite{PAWDQ}, which is made of 3D-printed plastic, has achieved a velocity of 1 m/s while walking . 
 
In this paper, we demonstrate the design and insights of a novel open-source quadruped robot HyperDog, which allows researchers and engineers both to tune and validate advanced locomotion algorithms in the Gazebo simulation environment based on ROS2 and to test these algorithms on a real quadruped robot HyperDog.


\section{HyperDog System Overview}
HyperDog is a 12-DoF legged robot with : 300 mm x 175 mm x 240 mm external dimension (WxHxD). Each leg of the robot consists of 3 joints for hip, upper and lower leg. The robot itself has a 5 kg weight and a 2 kg pay-load capability, and is powered by the 8.4V and 8.8Ah Li-Ion battery pack, which allows the HyperDog to run for about 50 minutes on a charge.

\begin{figure}[htbp]
\centering
\includegraphics[scale=0.23]{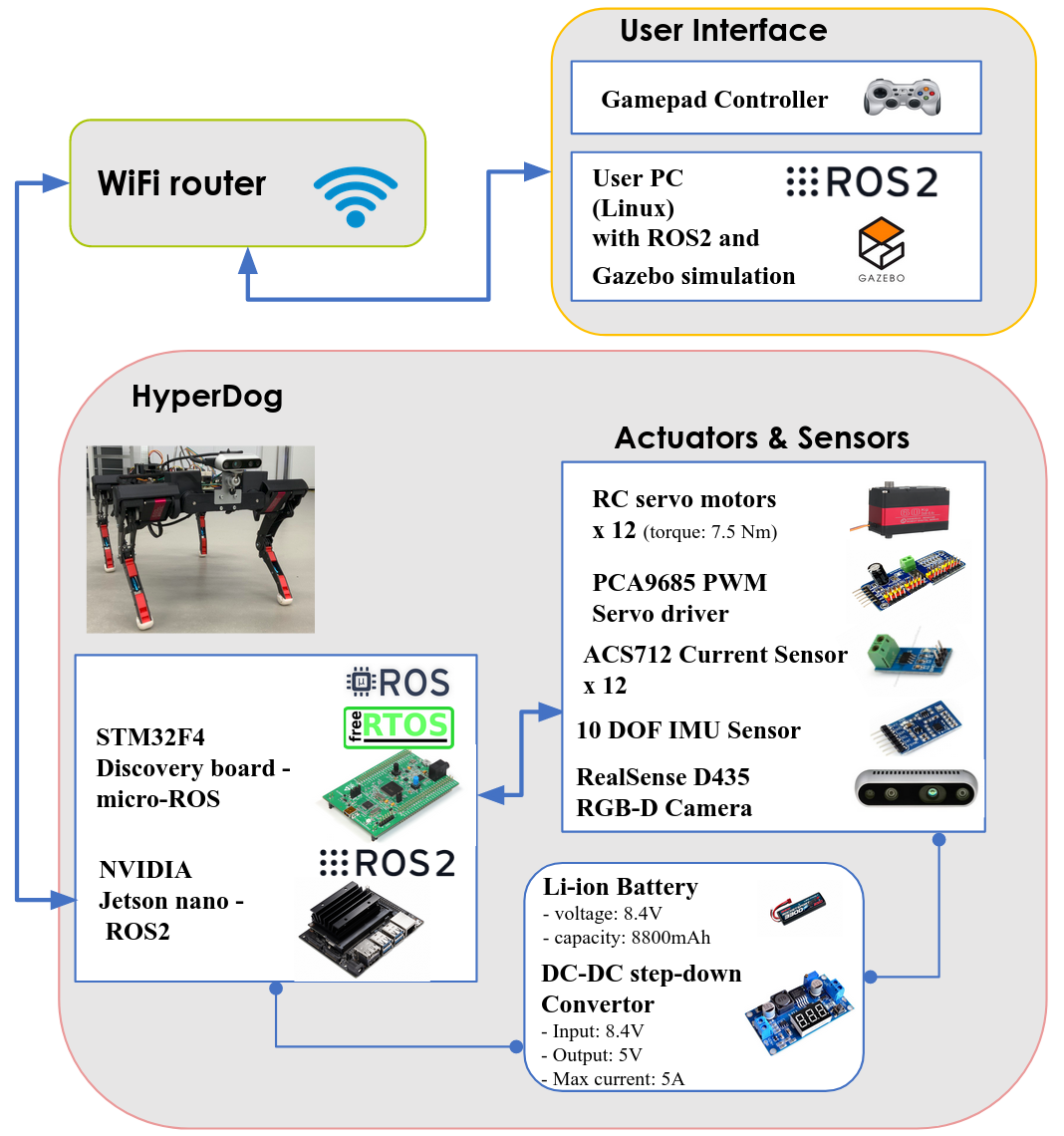}
\caption{HyperDog system overview.}
\label{fig:Overview}
\end{figure}

The developed HyperDog system consists of two main modules: robot interface and user interface (Fig. \ref{fig:Overview}). 

Robot interface includes onboard micro-controller STM32F4, computer NVIDIA Jetson nano, and actuators. The actuators include 12 radio control (RC) servo motors, an I2C-bus, 16-channel PWM Servo driver, 12 current sensors, and a 10 DoF IMU sensor. The communication between the onboard computer and the micro-controller is based on micro-ROS, which allows to establish fast and reliable data transmission. The user interface consists of a gamepad for HyperDog control and the personal computer (PC) with Intel Core i7 processor and Ubuntu OS, on which the simulation is conducted in the Gazebo environment. The PC and the main robot computer transfer information via Wi-Fi.

\subsection{Mechanical Design of the Leg}\label{AA}
HyperDog is equipped with the developed unique customizable legs. To reduce the weight of the leg structure, they were designed and produced from 3D printed materials and carbon fiber parts. The manufactured robotic legs are beneficial not only in regard to the lightweight structure but also due to their high strength and durability. The design of the leg plays a crucial role in the overall performance of the quadruped robot. Mass distribution, weight, capable payload, degrees of freedom, and workspace were considered when designing the leg for better performance. The mechanical specification of a designed leg is shown in Table \ref{tab:mechSpecs}.
\newcolumntype{M}[1]{>{\centering\arraybackslash}m{#1}}
\begin{table}[h!]
 \centering
 \caption{Mechanical specifications of the designed HyperDog's leg.}
 \begin{tabular}{ |M{4cm} | M{3cm}| }
 \hline
 \centering
 Technical criteria & Value \\
 \hline
 Degrees of freedom & 3 \\
 Hip link length & 104 mm \\ 
 Upper leg link length & 150 mm \\
 Lower leg link length & 150 mm \\
 Achievable minimum height & 80 mm \\
 Achievable maximum height & 240 mm \\
 Threshold value of damper & 148 N \\
 Displacement ratio of damper & 0.1 mm/N \\
 Maximum displacement of damper & 10 mm \\
 Weight & 0.95 kg \\
 Materials & PLA, TPU, Carbon Fiber\\
 \hline 
 \end{tabular}
 
 \label{tab:mechSpecs}
\end{table}

\begin{figure}[htbp]
\centering
\includegraphics[width=0.8\linewidth]{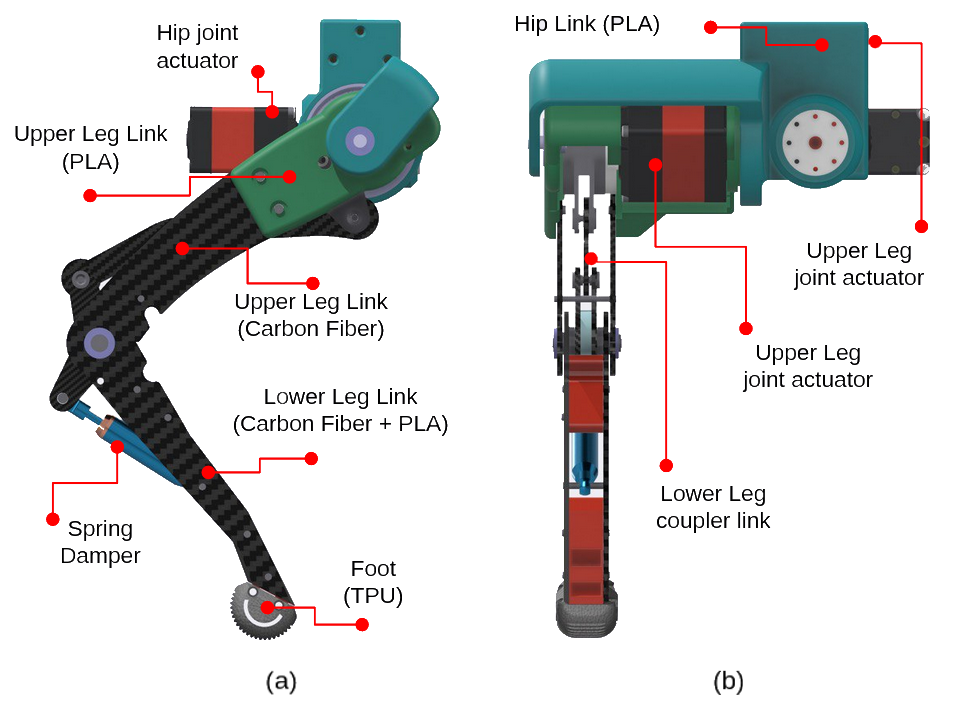}
\caption{The side (a) and front (b) views of the HyperDog's leg design. The lower and upper leg parts are designed from carbon fiber to reduce weight and increase durability, while hip joint with higher mass causing leg to have a low inertia during motion.}
\label{fig:legs}
\end{figure}

The spring damper is embedded in the lower leg to reduce the large impacts during the contact of robot's paw with the ground. The paw of the leg is designed and 3D printed using the thermoplastic polyurethane (TPU) material, which is flexible and strong enough to walk on harsh terrains. The developed leg is shown in Fig. \ref{fig:legs}. 

\begin{figure}[htbp]
\centering
\includegraphics[width=0.9\linewidth]{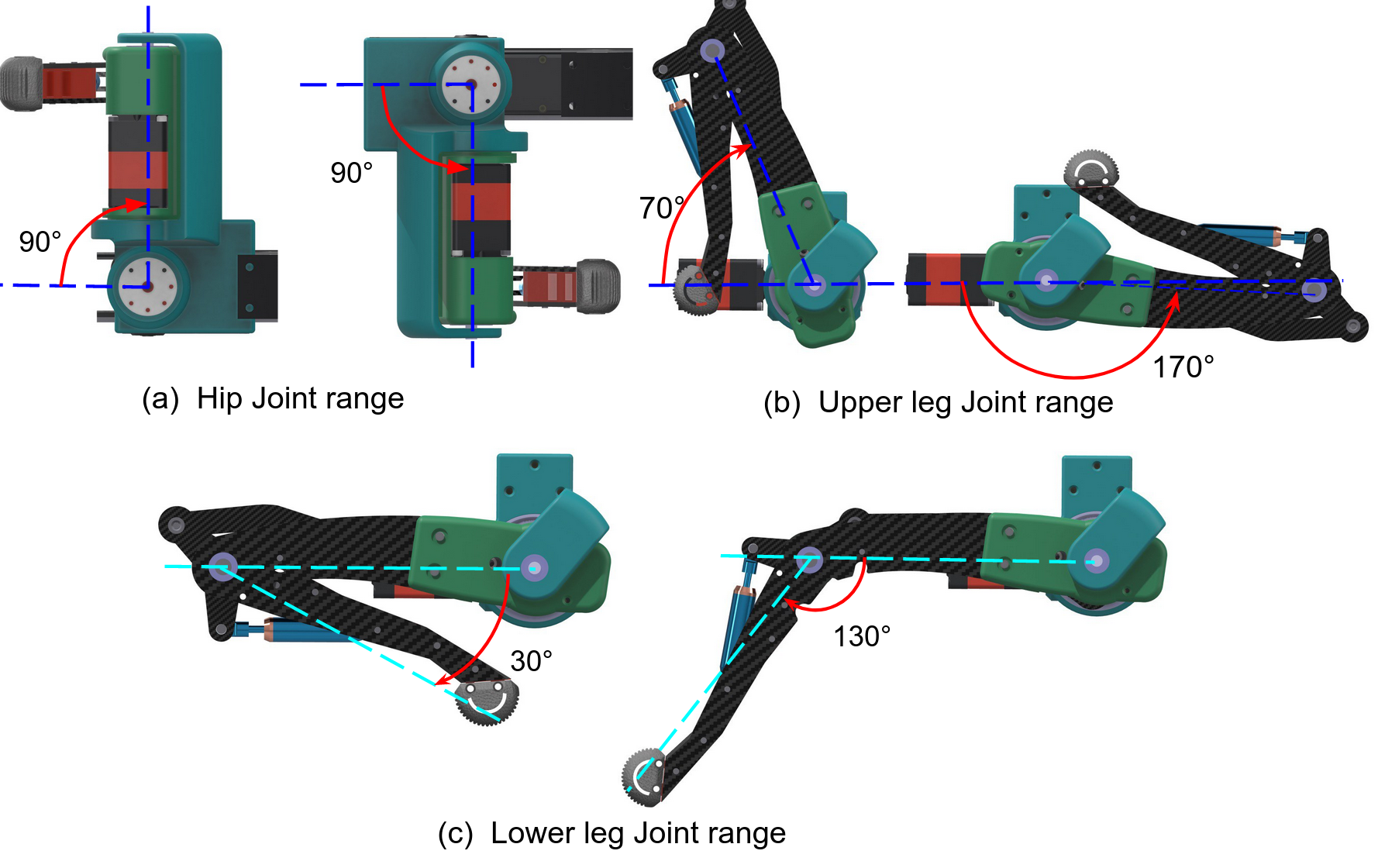}
\caption{Hip (a), Upper leg (b) and Lower leg (c) joint angle ranges of the designed HyperDog's leg. High range of motion is preserved for each joint, allowing robot to perform various gates while traversing terrain.}
\label{fig:jointRange}
\end{figure}

Each leg has three degrees of freedom: hip joint, upper leg joint, and lower leg joint. All the servo motors, which actuate the joints of the legs, are installed near the origin of the leg. That allows each leg to have low inertia during locomotion. Having low inertia is critical for the robot, as it has to be stable and perform locomotion algorithms efficiently. Moreover, it increases the stability of the platform, as the center of mass and geometrical center of the body are located closely. The joint limits of HyperDog's legs are shown in Fig. \ref{fig:jointRange}. 

The resulting range of motion for each joint of the HyperDog's leg is: -90$^{\circ}$ to 90$^{\circ}$ for the hip joint allowing 180$^{\circ}$ range, -70$^{\circ}$ to 170$^{\circ}$ for the upper leg joint allowing 240$^{\circ}$ range, and 30$^{\circ}$ to 130$^{\circ}$ for lower leg joint allowing 100$^{\circ}$ range of motion.

\subsection{Onboard Electronics}
\begin{figure}[h]
 \centering
 \includegraphics[width=0.8\linewidth]{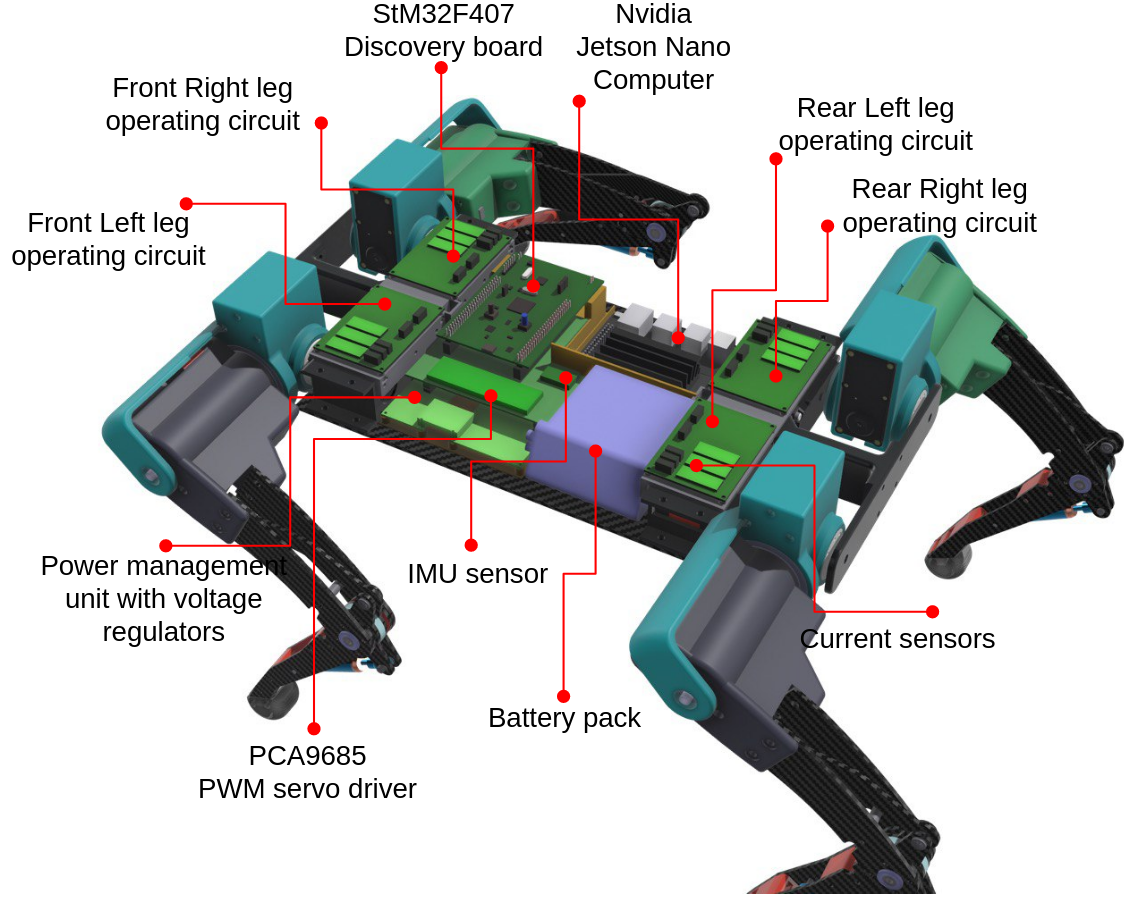}
 \caption{Onboard electronics of the HyperDog. }
 \label{fig:OnboardElectronics}
\end{figure}

As shown in Figure \ref{fig:OnboardElectronics}, the HyperDog robot is equipped with several control schemes, such as an onboard computer, microcontrollers, voltage regulators, current sensors, inertial measurement unit (IMU), and battery pack.

Ubuntu 20.04 was installed on the NVIDIA Jetson Nano computer, which is utilized as the main control unit of the robot. STM32F407 microcontroller-based Discovery board is used for controlling servo motors through the PCA9685 PWM servo driver. The STM32 microcontroller collects the sensor data from IMU, joint angle encoders, and current sensors. IMU is used to detect the orientation of the robot body, sudden accelerations and changes of direction caused by external disturbances. Current sensors are utilized for each servo motor to detect the intake current, after which this data is applied to recognize the contact between each foot and the ground and to calculate the applied force on each leg.
Joints are actuated by RDS5160 SSG high torque digital servo motors with 7 Nm maximum torque. Each servo motor is driven with 8.4 V and 2.5 A maximum current and has 150 g weight.

\subsection{Inverse Kinematic Model for Robot Legs and Body}
\begin{figure}
 \centering
 \begin{tabular}{@{}c@{}}

 \includegraphics[width=0.6\linewidth]{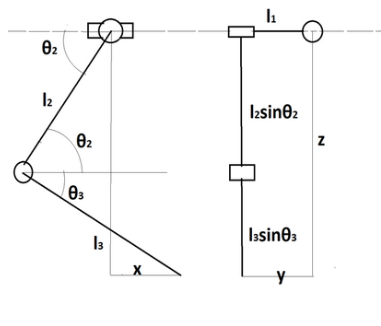} \\[\abovecaptionskip]
 \label{fig:leg_1}
 \vspace{-0.5em}
 \small (a) 
 \end{tabular}

 \vspace{0.5em}

 \begin{tabular}{@{}c@{}}
 \includegraphics[width=0.8\linewidth]{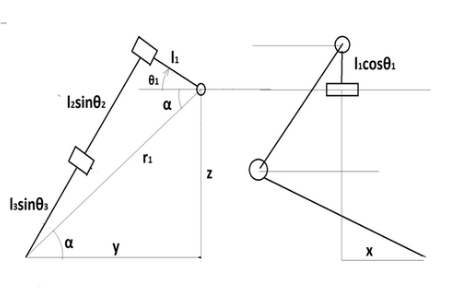} \\[\abovecaptionskip] \vspace{-0.5em}
 \label{fig:leg_2}
 
 \small (b) 
 \end{tabular}

 \caption{Link diagram of the leg when (a) $\theta_1 = 0$, (b) $\theta_1 > 0 $.}\label{fig:myfig}
\end{figure}

The inverse kinematics model is used to calculate the joint positions, which are needed to place the robot's end effector at a specific point and orientation. In order to move in 3D space, the robot leg should have three degrees of freedom. Therefore, the whole robot body has 12 degrees of freedom and can perform movements such as pitch, roll, and yaw.
The leg position is presented in Fig. \ref{fig:leg_1} - \ref{fig:leg_2}, where $l_1$ and $l_2$ are the link lengths, $\theta_1$, $\theta_2$ and $\theta_3$ are the hip, upper and lower leg joint angles respectively. 
The inverse kinematic solution for joint angles can be derived taking into account the leg position at $\theta_1 > 0$ (Fig. \ref{fig:leg_2}). 

\begin{figure}[htbp]
\centering
\includegraphics[scale=0.8]{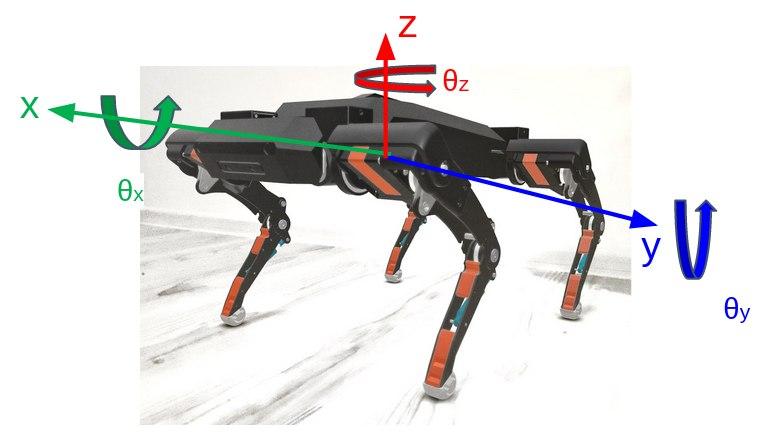}
\caption{The main robot coordinate system and pitch, roll and yaw angles.}
\label{fig: Kinem}
\end{figure}

The leg joint angles $\theta_1$, $\theta_2$, and $\theta_3$ are calculated as follows:

\begin{equation}
\theta_1 = \arccos(\frac{L_1}{r_1} - \alpha)
\label{eq_4}
\end{equation}

\begin{equation}
\theta_2 = \arcsin(\frac{c}{r_2} - \beta)
\label{eq_5}
\end{equation}

\begin{equation}
\theta_3 = \arcsin\Big(\frac{x + l_2 \cos(\theta_2)}{l_3} - \beta\Big)
\label{eq_6}
\end{equation}

\noindent
where $l_1$ and $l_2$ are the upper and lower link lengths, $r_1$ and $r_2$ are the sides of right triangles that are derived in the side (XZ) and top (XY) projections of the leg, $\beta$ is the $\arctan2(a, b)$, and $a, b, c$ are the leg length parameters expressed as follows:

\begin{equation}
a = 2\cdot l_2 \cdot r_1 \cdot \sin(\theta_1 + \alpha)
\label{eq_1}
\end{equation}

\begin{equation}
b = 2\cdot x\cdot l_2
\label{eq_2}
\end{equation}

\begin{equation}
c = x^2 + l_2^2 - l_3^2 + (\sin(\theta_1 + \alpha))^2 
\label{eq_3}
\end{equation}

\noindent
where $x$ is the length of the robot's step.
In order to achieve free HyperDog movements, the robot is able to rotate its body around X, Y and Z axis without any change in paw position, as shown in Fig. \ref{fig: Kinem}.

For this purpose rotation matrices are calculated as given as in:

\begin{equation}
 R_x = \begin{bmatrix}
 1 & 0 & 0 \\
 0 & \cos(\theta_x) & -\sin(\theta_x) \\
 0 & \sin(\theta_x) & \cos(\theta_x)
 \end{bmatrix}
 \label{eq_7}
\end{equation}

\begin{equation}
 R_y = 
 \begin{bmatrix}
 \cos(\theta_y) & 0 & \sin(\theta_y) \\
 0 & 1 & 0\\
 -\sin(\theta_x) & 0 & \cos(\theta_y)
 \end{bmatrix}
\label{eq_8}
\end{equation}

\begin{equation}
 R_z = \begin{bmatrix}
 \cos(\theta_z)& -\sin(\theta_z) & 0 \\
 \sin(\theta_z)& \cos(\theta_z) & 0 \\
 0 & 0 & 1
 \end{bmatrix}
 \label{eq_9}
\end{equation}

\noindent
where $\theta_x$, $\theta_y$, and $\theta_z$ are the rotation axis-angles of the HyperDog body.
Then the rotation matrix is calculated by multiplying rotation matrices around each axis, as given in: 

\begin{equation}
 R(\theta_z, \theta_y, \theta_x) = R_z \cdot R_y \cdot R_x
\label{eq_10}
\end{equation}

From this rotation matrix the paw coordinates relative to the body center are derived for each leg.

\subsection{ROS2 HyperDog Control Packages}

The HyperDog is entirely controlled with the help of ROS2 and micro-ROS packages allowing researchers and other developers to launch their own locomotion algorithms within the HyperDog platform and to debug quite easily. Micro-ROS is utilized for HyperDog's low-level algorithms, which are running on the STM32F4 Discovery board.
In fact, micro-ROS is the robotic framework that bridges the gap between resource-constrained and larger processing units and robotic applications. It brings the ROS programming interfaces to these resource-constrained devices. With micro-ROS, microcontrollers can hereby be integrated seamlessly into ROS2-based systems, therefore no longer being inflexible black boxes and bringing all the benefits of ROS technology into robotics product development. 
Hereby, micro-ROS can subscribe to and publish ROS2 topics with the help of a micro-ROS agent, which is launched on the onboard computer NVIDIA Jetson Nano.

\begin{figure}[htbp]
\centering
\includegraphics[scale=0.2]{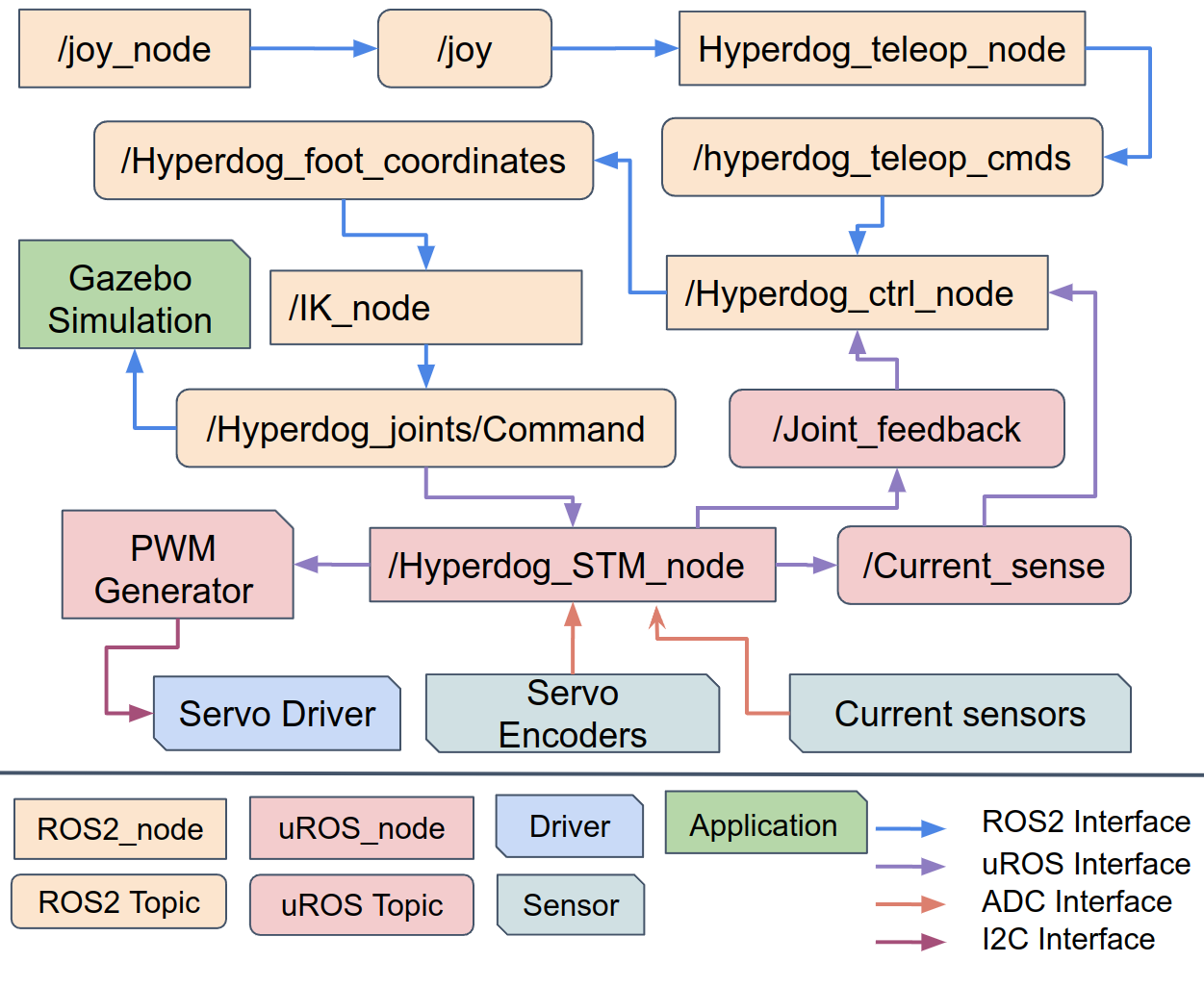}
\caption{ROS2 and micro-ROS based HyperDog control pipeline.}
\label{fig:ROS}
\end{figure}

ROS2 packages can be run on a PC or robot's onboard NVIDIA Jetson Nano computer. Developed ROS2 packages for controlling the HyperDog as following: Hyperdog\_teleop\_pkg, Hyperdog\_ctrl\_pkg, HyperDog\_joint\_ctrl\_pkg and Hyperdog\_gazebo\_simulation. Moreover, the Joy ROS2 open source package for reading buttons and joystick states of the Gamepad are used for controlling the HyperDog. The whole ROS2 and micro-ROS-based HyperDog control pipeline (Fig. \ref{fig:ROS}) can be described in the following way.

\begin{figure}[htbp]
\centering
\includegraphics[scale=0.18]{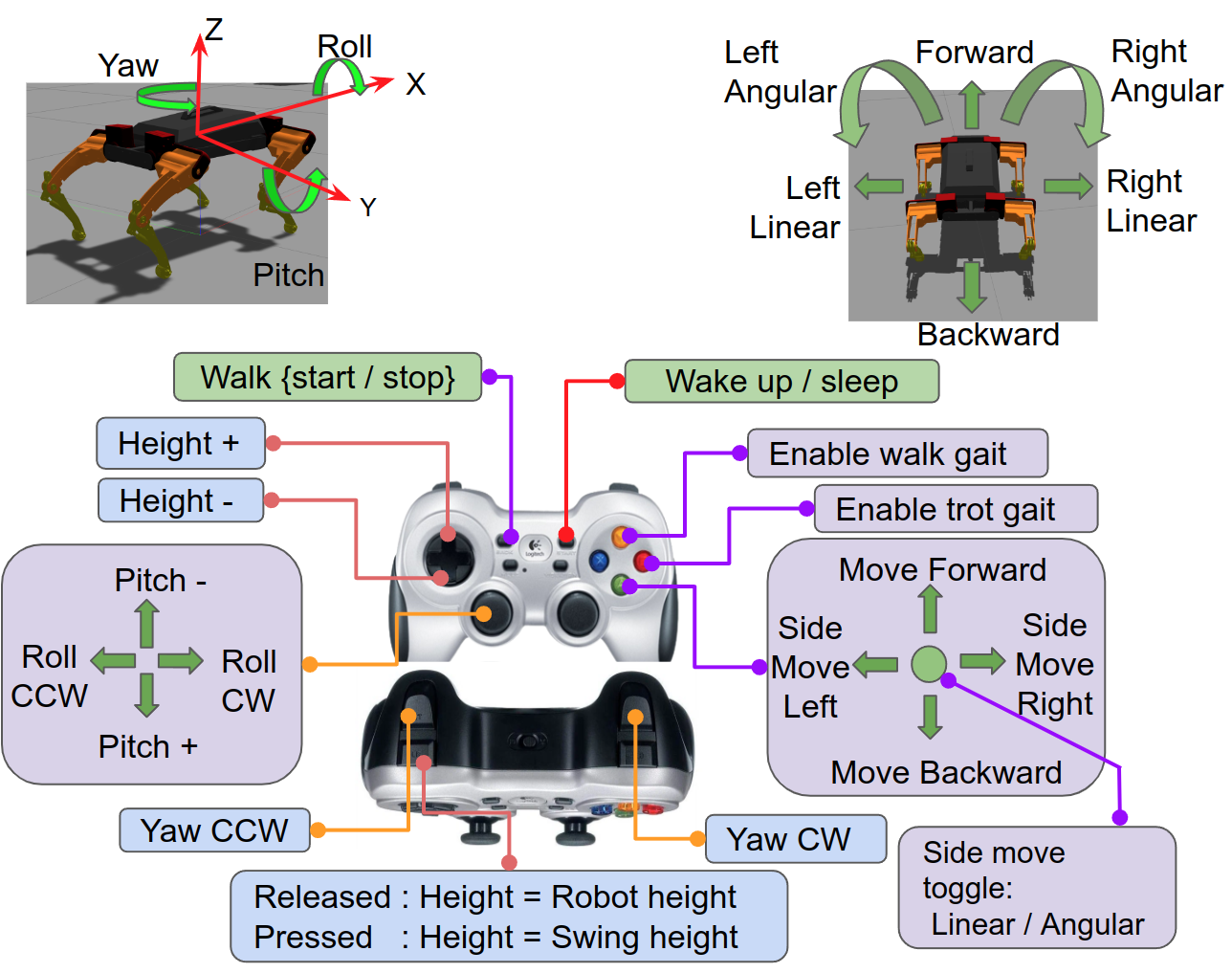}
\caption{Control commands from joystick gamepad.}
\label{fig:JoyCmds}
\end{figure}

/Joy\_node reads the Gamepad buttons and joysticks states and then publishes this information through /joy topic. Developed HyperDog\_teleop package 
subscribes to /joy topic and makes commands which are: start, walk, side\_walk\_mode\{linear/rotation\}, gait\_pattern, step\_length\{x,y\}, step\_height\{swing, stance\}, robot\_height, euler\_angles\{roll, pitch, yow\} (Fig. \ref{fig:JoyCmds}). 

These commands are then published through /Hyperdog\_teleop\_cmds which is subscribed to /Hyperdog\_ctrl\_node. In turn, /Hyperdog\_ctrl\_node is run by the Hyperdog\_ctrl package and it has four main tasks such as, reading commands, generating gait, whole body controlling, and solving inverse kinematics. /Hyperdog\_ctrl\_node uses parallel processing for reading commands, gait generating, and whole body controlling to reduce the reaction time delay, and then it publishes the end effector coordinate for each leg through /Hyperdog\_foot\_coordinate. As for inverse kinematics, it is solved by /IK\_node, which subscribes to Hyperdog\_foot\_coordinates, and then calculates angle values for each joint of all the legs. Next, it publishes through /Hyperdog\_joints/Commands topic, which can be subscribed by joint\_controller of Hyperdog\_gazebo\_simulation package to operate and test algorithms in the simulation. Moreover, with the help of micro-ROS agent /Hyperdog\_joints/Commands topic running on the onboard computer is subscribed by /Hyperdog\_STM\_node, which is a micro-ROS node running on STM32F407 microcontroller. The micro-controller controls servo motors with the help of PCA9685 PWM driver, reads and processes servo motors encoder data and current sensor data, which are then published through /Joint\_feedback and /Current\_sense topics to /Hyperdog\_ctrl\_node.

\begin{figure}[htbp]
\centering
\includegraphics[scale=0.33]{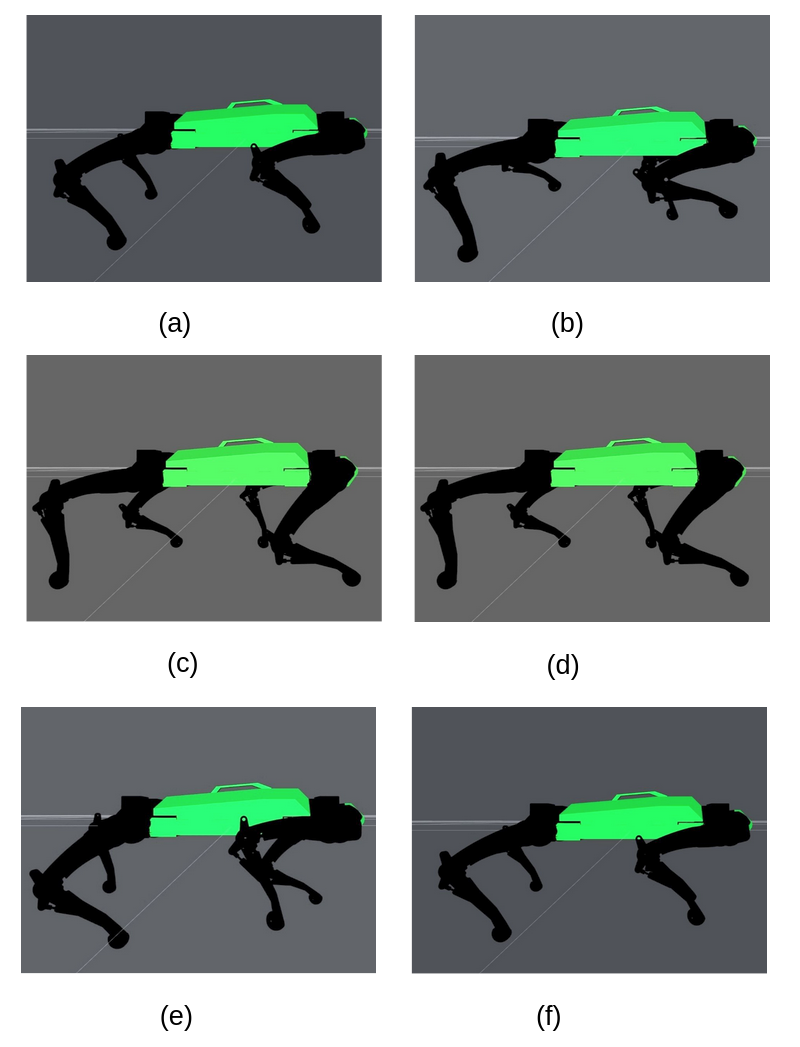}
\caption{Trotting gait steps. (a) All legs are on the ground. (b) Back Left (BL) and Front Right (FR) legs are in a swing while Back Right (BR) and Front Left (FL) legs are in a stance. (c) BL and FR legs complete the swing while BR and FL legs complete the stance. (d) BL and FR legs starting a swing while BR and FL starting a stance. (e) BL and FR legs in the swing, while BR and FL legs are in the stance. (f) All legs are on the ground.}
\label{fig:trot}
\end{figure}

We have provided a separate folder in the Hyperdog\_ctrl package with necessary python scripts for the Gait Planner, where your gait algorithms can be tested. Here we will demonstrate two examples of trotting gait and walking gait algorithms which we have tested in the simulation environment. Trotting gait is shown in Fig. \ref{fig:trot}.

\begin{figure}[htbp]
\centering
\includegraphics[scale=0.35]{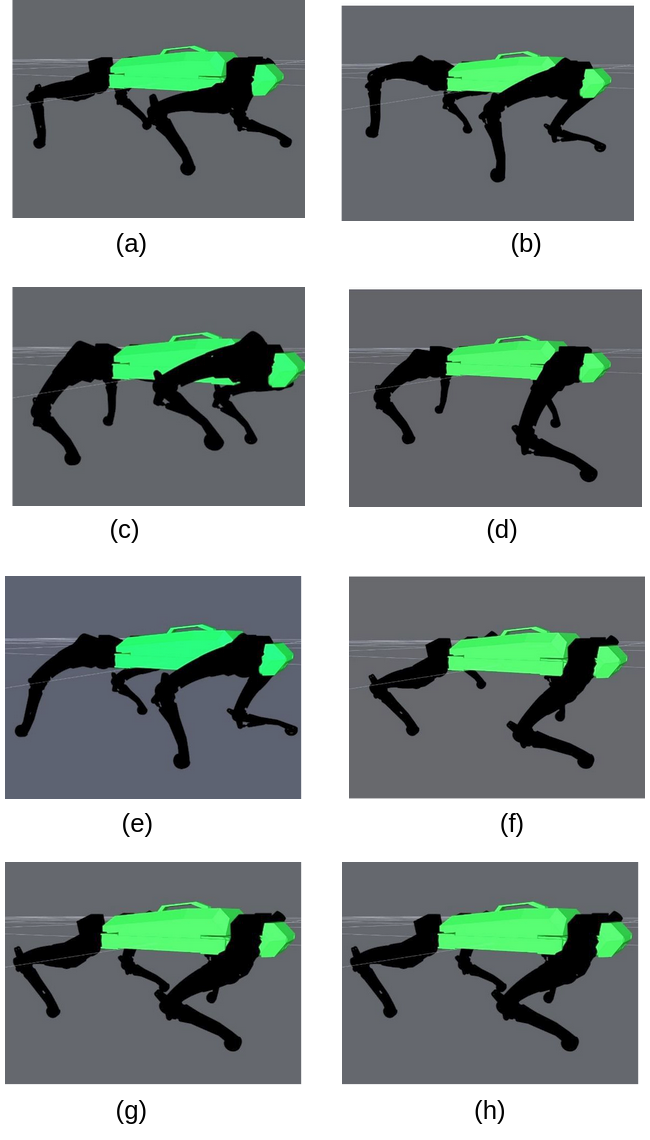}
\caption{Walking gait steps. (a) Body leans to Left side. (b) BR leg is in a swing while other legs stance. (c) BR leg completes the swing and FR leg starts a swing while other legs stance. (d) FR leg completes the swing. (e) Body leans to the right side. (f) BL leg swings while other legs in a stance. (g) BL completes the swing, FL starts to swing, and other legs are in a stance. (h) FL leg completes the swing and body leans to the left side. }
\label{fig:walk}
\end{figure}

The walking gait is shown in Fig. \ref{fig:walk}. In the gait the body is leaned to keep the center of mass inside the triangle when the other legs make contact with the ground.


\subsection{HyperDog Gazebo Simulation Environment}

\begin{figure}[htbp]
\centering
\includegraphics[scale=0.12]{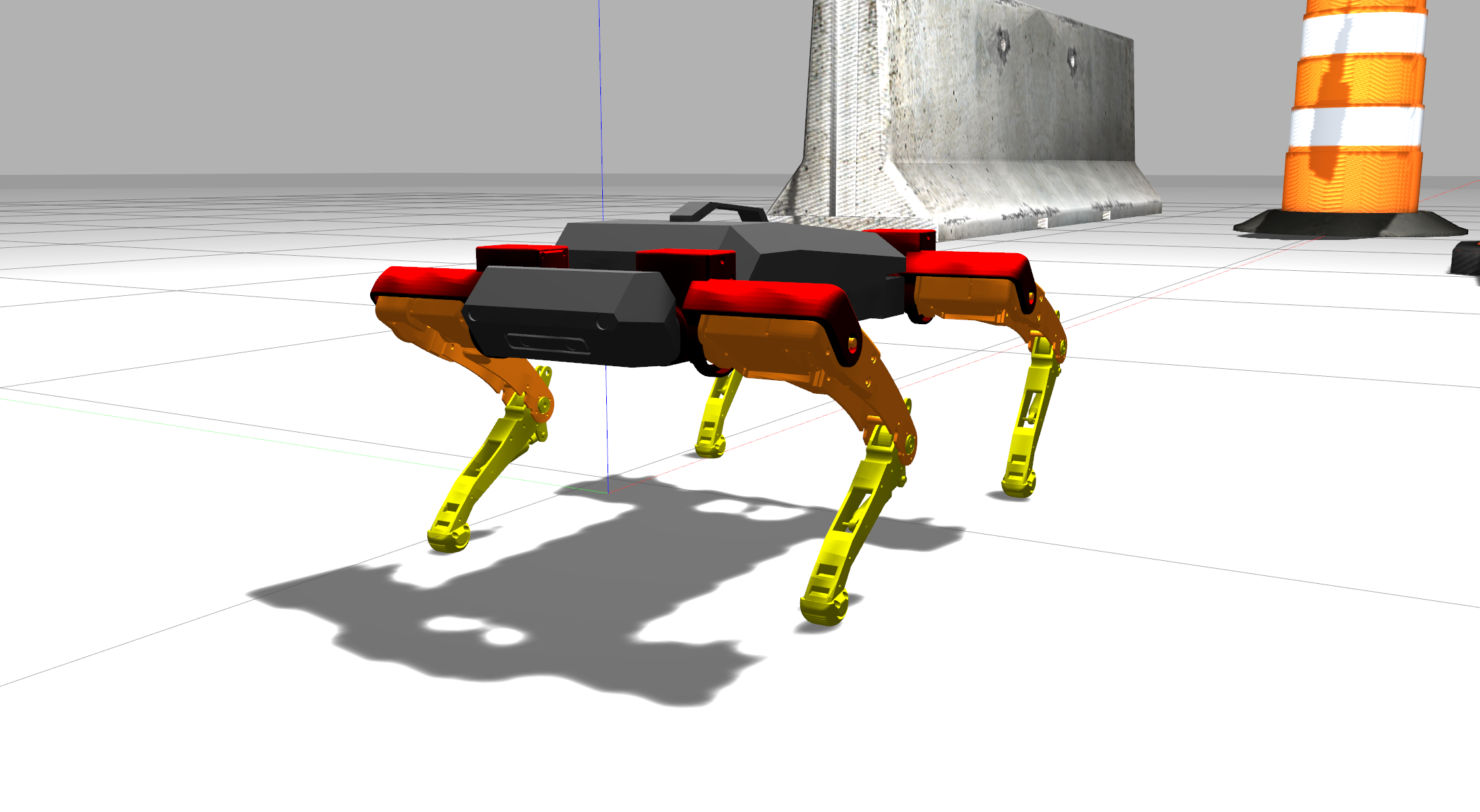}
\caption{HyperDog simulation environment in Gazebo.}
\label{fig:Gazebo}
\end{figure}

The HyperDog simulation environment is shown in Fig. \ref{fig:Gazebo}. As mentioned before, the developed HyperDog platform allows researchers and developers to simulate the HyperDog in a Gazebo simulation environment. This simulation environment helps researchers and developers to run and test their own algorithms on the HyperDog platform without harming the physical robot. As well as, the HyperDog simulation environment can be used as a digital twin of the HyperDog since the same control package is used by the physical robot and simulation. Moreover, with this simulation package, one can simulate the robot in various worlds and add additional obstacles and objects making the terrain denser and harsher. The simulation is developed for the Gazebo-11 version and it can be easily launched by running the terminal command \emph{"ros2 launch hyperdog\_gazebo\_sim hyperdog\_gazebo\_sim.launch.py"} on Ubuntu with the installed HyperDog package in ROS2 environment. Once the simulation is opened it subscribes to the ROS2 topic /Hyperdog\_joints/Commands of Hyperdog\_ctrl\_node and simulates the joint angles according to given commands from the gamepad.





\section{Conclusion and Future Work}
This work proposes a novel open-source quadruped platform HyperDog, which gives a possibility to researchers and engineers to tune and test locomotion algorithms on the HyperDog model in the simulation environment and on the real robot HyperDog, which we both have developed. All the control system algorithms were designed within the open-source robotic framework
ROS2 and Micro-ROS allow further developments to be easily implemented.

In the future, the plan is to apply a dynamic model and extend the HyperDog platform by adding the ability to perform control analysis. Moreover, we plan to implement visual localization using more RGB-D and infrared cameras to monitor the environment ahead and improve locomotion planners to overcome obstacles.



\bibliographystyle{IEEEtran}
\bibliography{cites}

\end{document}